\documentclass{article}

\PassOptionsToPackage{numbers, compress}{natbib}
\usepackage[final]{nips_2017}

\usepackage[utf8]{inputenc} 
\usepackage[T1]{fontenc}    
\usepackage{hyperref}       
\usepackage{url}            
\usepackage{booktabs}       
\usepackage{amsfonts}       
\usepackage{nicefrac}       
\usepackage{microtype}      
\usepackage{amsmath}
\usepackage{graphicx}
\newcommand{\norm}[1]{\left\lVert#1\right\rVert}
\newcommand{\E}[2]{\mathbb{E}_{#1}\left[#2\right]}
\newcommand{\Prob}[2]{\mathbb{P}\left[#1 \mid #2\right]}
\newcommand{\R}{\mathbb{R}}
\title{Adversarial Patch}

\author{Tom B. Brown, Dandelion Man\'e\thanks{ Equal contributions with Tom B. Brown }, Aurko Roy, Mart\'in Abadi, Justin Gilmer  \\
\texttt{\{tombrown,dandelion,aurkor,abadi,gilmer\}@google.com}
}

\begin{document}

\maketitle

\begin{abstract}
We present a method to create universal, robust, targeted adversarial image patches in the real world. The patches are universal because they can be used to attack any scene, robust because they work under a wide variety of transformations, and targeted because they can cause a classifier to output any target class. These adversarial patches can be printed, added to any scene, photographed, and presented to image classifiers; even when the patches are small, they cause the classifiers to ignore the other items in the scene and report a chosen target class. 
\end{abstract}

\section{Introduction}
\label{sec:introduction}

Deep learning systems are broadly vulnerable to adversarial examples, 
carefully chosen inputs that cause the network to change output without a visible change to a human \citep{Szegedy14, 
goodfellow2014explaining}. These adversarial examples most commonly modify 
each pixel by only a small amount and can be found using a number of 
optimization strategies such as L-BFGS \citep{Szegedy14}, Fast Gradient Sign 
Method (FGSM) \citep{goodfellow2014explaining}, DeepFool 
\citep{moosavi2016deepfool}, Projected Gradient Descent (PGD) 
\citep{madry2017advexamples}, as well as the recently proposed 
Logit-space Projected Gradient Ascent (LS-PGA) 
\cite{anonymous2018thermometer} for discretized inputs. 
Other attack methods seek to modify only a 
small number of pixels in the image (Jacobian-based saliency map 
\citep{papernot2016limitations}), or a small patch at a fixed location of the image 
\citep{sharif2016accessorize}.

Adversarial examples have been shown to generalize to the real world. Kurakin et al.~\cite{kurakin2016adversarial} demonstrated that when printed out, an adversarially constructed image will continue to be adversarial to classifiers even under different lighting and orientations. Athalye et al.~\cite{2017synthesizing} recently demonstrated adversarial objects which can be 3d printed and misclassified by networks at different orientations and scales. Their adversarial objects are designed to be subtle perturbations of a normal object (e.g. a turtle that has been adversarially perturbed to be classified as a rifle). Another work~\citep{sharif2016accessorize} showed that one can fool facial recognition software by constructing adversarial glasses. These glasses were targeted in that they could be constructed to impersonate any person, but were custom made for the attacker's face, and were designed with a fixed orientation in mind. Even more recently, Evtimov et al.~\citep{evtimov2017robust} demonstrated various methods for constructing stop signs that are misclassified by models, either by printing out a large poster that looks like a stop sign, or by placing various stickers on a stop sign. In terms of defenses there has been substantial work on increasing the adversarial robustness of image models to small $L_p$ perturbations of the input \cite{madry2017advexamples,papernot2016distillation, tramer2017ensemble, anonymous2018thermometer}.

As seen above, a majority of prior work has focused on attacking with and defending against either small or imperceptible changes to the input. In this work we explore what is possible if an attacker no longer restricts themselves to imperceptible changes. We construct an attack that does not attempt to subtly transform an existing item into another. Instead, this attack generates an image-independent patch that is extremely salient to a neural network. This patch can then be placed anywhere within the field of view of the classifier, and causes the classifier to output a targeted class. Because this patch is scene-independent, it allows attackers to create a physical-world attack without prior knowledge of the lighting conditions, camera angle, type of classifier being attacked, or even the other items within the scene. 

\begin{figure}[h]
  \label{fig:real-world-attack}
  \centering
  \includegraphics[width=1.0\linewidth]{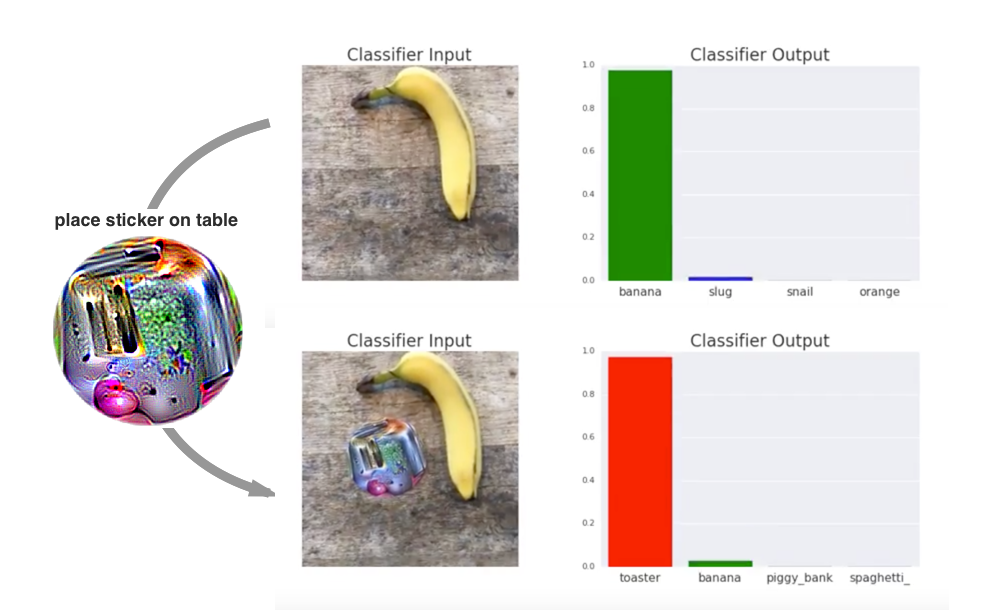}
  \caption{A real-world attack on VGG16, using a physical patch generated by the white-box ensemble method described in Section 3. When a photo of a tabletop with a banana and a notebook (top photograph) is passed through VGG16, the network reports class 'banana' with 97\% confidence (top plot). If we physically place a sticker targeted to the class "toaster" on the table (bottom photograph), the photograph is classified as a toaster with 99\% confidence (bottom plot). See the following video for a full demonstration: \url{https://youtu.be/i1sp4X57TL4}
}
\end{figure}

This attack is significant because the attacker does not need to know what image they are attacking when constructing the attack. After generating an adversarial patch, the patch could be widely distributed across the Internet for other attackers to print out and use. Additionally, because the attack uses a large perturbation, the existing defense techniques which focus on defending against small perturbations may not be robust to larger perturbations such as these. Indeed recent work has demonstrated that state-of-the art adversarially trained models on MNIST are still vulnerable to larger perturbations than those used in training either by searching for a nearby adversarial example using a different metric for distance \citep{chen2017madry}, or by applying large perturbations in the background \citep{anonymous2018adversarial}.

\section{Approach}
\label{sec:approach}

The traditional strategy for finding a targeted adversarial example is as follows: given some classifier \(\Prob{y}{x}\), some input \(x \in \R^n\), some target class \(\widehat{y}\) 
and a maximum perturbation \(\varepsilon\), we want to find the input \(\widehat{x}\) that maximizes 
\(\log \left(\Prob{\widehat{y}}{\widehat{x}}\right)\), subject to the constraint that \(\norm{x-\widehat{x}}_\infty \leq \varepsilon\). When \(\Prob{y}{x}\) is parameterized 
by a neural network, an attacker with access to the model can perform iterated gradient descent on \(x\) in order to find a suitable input \(\widehat{x}\). This strategy can produce a well camouflaged attack, but requires modifying the target image. 

\begin{figure}[h]
  \label{fig:patch-op}
  \centering
  \includegraphics[width=1.0\linewidth]{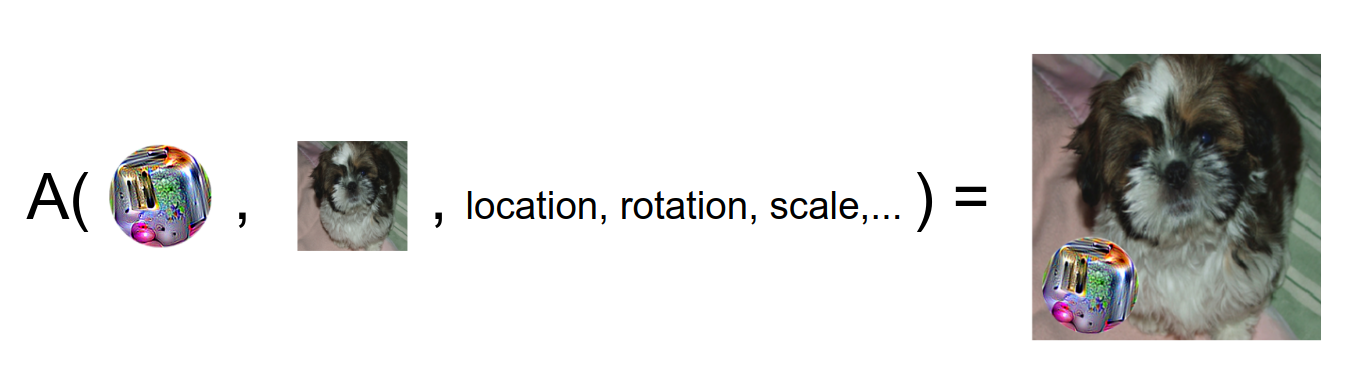}
  \caption{An illustration of the patch application operator. The operator takes as input a patch, an image, a location, and any patch transformations (such as scale and rotations) and applies the transformed patch to the image at the given location. The patch is then trained to optimize the expected probability of a target class, where the expectation is taken over random images, locations, and transformations.}
\end{figure}

Instead, we create our attack by completely replacing a part of the image with our patch. We mask our patch to allow it to take any shape, and then train over a variety of images, applying a random translation, scaling, and rotation on the patch in each image, optimizing using gradient descent. In particular for a given image $x \in \mathbb{R}^{w \times h \times c}$, patch $p$, patch location $l$, and patch transformations $t$ (e.g. rotations or scaling) we define a \emph{patch application operator} $A(p, x, l, t)$ which first applies the transformations $t$ to the patch $p$, and then applies the transformed patch $p$ to the image $x$ at location $l$ (see figure~\ref{fig:patch-op}). 

To obtain the trained patch $\widehat{p}$ we use a variant of the Expectation over Transformation (EOT) framework of Athalye et al. \citep{2017synthesizing}. In particular, the patch is trained to optimize the objective function

\begin{equation} \label{eq:patch_obj}
    \widehat{p} = \arg\max_{p}\E{x \sim X, t \sim T, l \sim L}{\log{\Pr(\widehat{y} | A(p, x, l, t)}}
\end{equation}

where $X$ is a training set of images, $T$ is a distribution over transformations of the patch, and $L$ is a distribution over locations in the image. Note that this expectation is over images, which encourages the trained patch to work regardless of what is in the background. This departs from most prior work on adversarial perturbations in the fact that this perturbation is \emph{universal} in that it works for any background. Universal perturbations were identified in \citep{moosavi2016universal}, but these required changing every pixel in the image and results were not given in the physical world.

We also consider camouflaged patches which are forced to look like a given starting image. Here we simply add a constraint of the form $||p - p_{orig}||_\infty < \epsilon$ to the patch objective. This will force the final patch to be within $\epsilon$ in the $L_\infty$ norm of some starting patch $p_{orig}$.

We believe that this attack exploits the way image classification tasks are constructed. While images may contain several items, only one target label is considered true, and thus the network must learn to detect the most "salient" item in the frame. The adversarial patch exploits this feature by producing inputs much more salient than objects in the real world. Thus, when attacking object detection or image segmentation models, we expect a targeted toaster patch to be classified as a toaster, and not to affect other portions of the image.

\section{Experimental Results}
\label{sec:results}

To test our attack, we compare the efficacy of two whitebox attacks, a blackbox attack, and a control patch. The white box ensemble attack jointly trains a single patch across five ImageNet models: inceptionv3, resnet50, xception, VGG16, and VGG19. We then evaluate the attack by averaging the win rate across all five models. The white box single model attack does the same but only trains and evaluates on a single model. The blackbox attack jointly trains a single patch across four of the ImageNet models, and then evaluates the blackbox attack on a fifth model, which we did not access during training. The control is a picture of a toaster.

\begin{figure}[h]
  \label{fig:attack-success}
  \centering
  \includegraphics[width=1.0\linewidth]{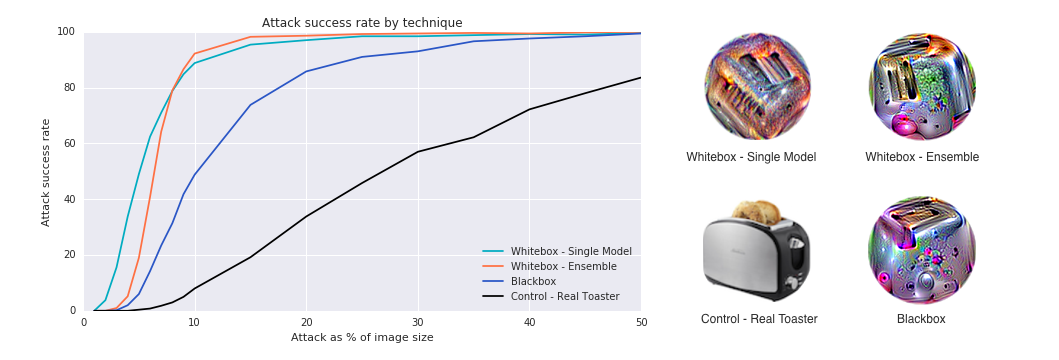}
  \caption{A comparison of different methods for creating adversarial patches. Note that these success rates are for random placements of the patch on top of the image. Each point in the plot is computed by applying the patch to 400 randomly chosen test images at random locations in these images. This is done for various scales of the patch as a fraction of the size of the image, each scale is tested independently on 400 images. 
}
\end{figure}

During training and evaluation, the patches are rescaled and then digitally inserted on a random location on a random ImageNet image. Figure 2 shows the results. 


Note that the patch size required to reliably fool the model in this universal setting (black box, on a targeted class, and over all images, locations and transformations) is significantly larger than those required to perform a non-targeted attack on a single image and a single location in the whitebox setting. For example, Su et al.~\citep{one_pixel} recently demonstrated that modifying 1 pixel on a 32x32 pixel CIFAR-10 image (0.1\% of the pixels in the image) suffices to fool the majority of images with a non-targeted, non-universal whitebox attack. However, our attacks are still far more effective than naively inserting an image with the target class, as is shown in Figure 2 by the relatively poor performance of inserting a real toaster into the scene. 

Any of the attacks shown in Figure 2 can be camouflaged in order to reduce their saliency to a human observer. We create a disguised version of the patch by minimizing its L2 distance to a tie-dye pattern and applying a peace sign mask during training. The results from these experiments are found in Figure 3.

\begin{figure}[h]
  \centering
    \includegraphics[width=1.0\linewidth]{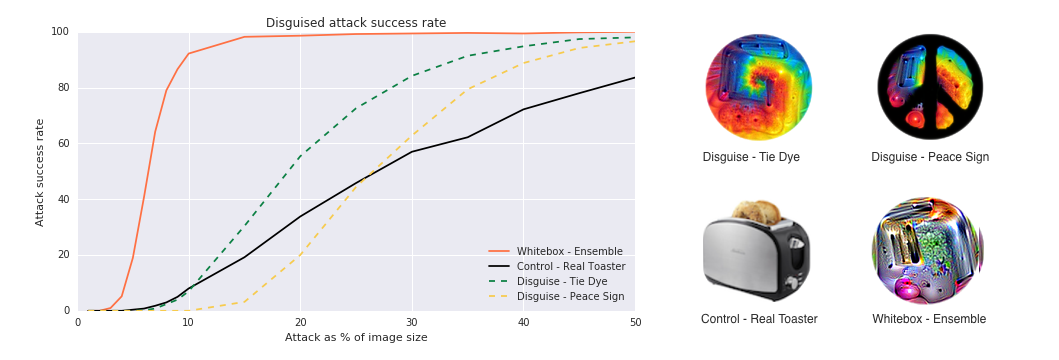}
  \caption{A comparison of patches with various disguises. We find that we can disguise the patch and retain much of its power to fool the classifier.
}
\end{figure}

In our final experiment, we test the transferability of our attack into the physical world. We print the generated patch with a standard color printer, and put it a variety of real world situations. The results shown in Figure 1 demonstrate that the attack successfully fools the classifier, even when there are other objects within the scene. For a full video demonstration of the attack, see \url{https://youtu.be/i1sp4X57TL4}.

We also tested the black box + physical world effectiveness of the patch on the third party Demitasse application\footnote{\url{https://itunes.apple.com/us/app/demitasse-image-recognition-cam/id1138211169?mt=8}} and found some transferability of the patch but only when the patch takes up a significant fraction of the image. We did not optimize the patch for print-ability as in \cite{sharif2016accessorize}, which perhaps explains why the patch is not as effective as in Figure~\ref{fig:attack-success}, which tests black box for different models and not in the physical world. We invite curious readers to try the patch out for themselves by printing out this paper and using the patch in the Appendix. 

\section{Conclusion}
\label{sec:conclusion}

We show that we can generate a universal, robust, targeted patch that fools classifiers regardless of the scale or location of the patch, and does not require knowledge of the other items in the scene that it is attacking. Our attack works in the real world, and can be disguised as an innocuous sticker. These results demonstrate an attack that could be created offline, and then broadly shared.

There has been substantial work on defending against small $L_p$ perturbations to natural images, at least partially motivated by security concerns \cite{papernot2016distillation,madry2017advexamples,anonymous2018thermometer}. Part of the motivation of this work is that potential malicious attackers may not be concerned with generating small or imperceptible perturbations to a natural image, but may instead opt for larger more effective but noticeable perturbations to the input - especially if a model has been designed to resist small $L_p$ perturbations.  

Many ML models operate without human validation of every input and thus malicious attackers may not be concerned with the imperceptibility of their attacks. Even if humans are able to notice these patches, they may not understand the intent of the patch and instead view it as a form of art. This work shows that focusing only on defending against small perturbations is insufficient, as large, local perturbations can also break classifiers.


\section{Appendix}
\label{sec:appendix}

\begin{figure}[h]
  \label{fig:printable-attack}
  \centering
  \includegraphics[width=0.8\linewidth]{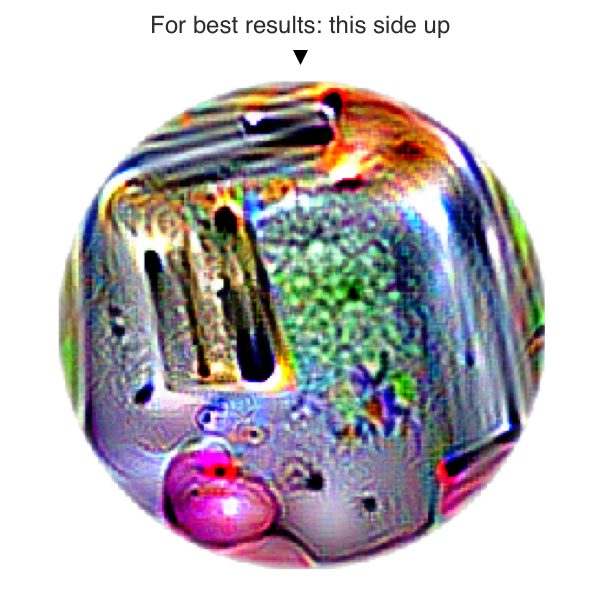}
  \vspace{10pt}
  \caption{Printable sticker illustrating the attack. For best results keep stickers within 20 degrees of the vertical alignment shown here. This patch was generated by the white-box ensemble method described in Section 3. We observed some transferability of this patch to the third party Demitasse application (the patch was not designed to fool this application). However, in order to be effective the size of the patch needs to be larger than what is demonstrated in Figure~\ref{fig:real-world-attack}, which is a white box attack on the models described in Section~\ref{sec:results}. 
}
\end{figure}

\bibliographystyle{abbrv}
\bibliography{patch}

\begin{thebibliography}{10}

\bibitem{anonymous2018adversarial}
Anonymous.
\newblock Adversarial spheres.
\newblock {\em International Conference on Learning Representations}, 2018.

\bibitem{anonymous2018thermometer}
Anonymous.
\newblock Thermometer encoding: One hot way to resist adversarial examples.
\newblock {\em International Conference on Learning Representations}, 2018.

\bibitem{2017synthesizing}
A.~Athalye, L.~Engstrom, A.~Ilyas, and K.~Kwok.
\newblock Synthesizing robust adversarial examples.
\newblock {\em arXiv preprint arXiv:1707.07397}, 2017.

\bibitem{evtimov2017robust}
I.~Evtimov, K.~Eykholt, E.~Fernandes, T.~Kohno, B.~Li, A.~Prakash, A.~Rahmati,
  and D.~Song.
\newblock Robust physical-world attacks on deep learning models.
\newblock {\em arXiv preprint arXiv:1707.08945}, 2017.

\bibitem{goodfellow2014explaining}
I.~J. Goodfellow, J.~Shlens, and C.~Szegedy.
\newblock Explaining and harnessing adversarial examples.
\newblock {\em arXiv preprint arXiv:1412.6572}, 2014.

\bibitem{one_pixel}
S.~K. Jiawei~Su, Danilo Vasconcellos~Vargas.
\newblock One pixel attack for fooling deep neural networks.
\newblock {\em arXiv preprint arXiv:1710.08864}, 2017.

\bibitem{kurakin2016adversarial}
A.~Kurakin, I.~Goodfellow, and S.~Bengio.
\newblock Adversarial examples in the physical world.
\newblock {\em arXiv preprint arXiv:1607.02533}, 2016.

\bibitem{madry2017advexamples}
A.~Madry, A.~Makelov, L.~Schmidt, D.~Tsipras, and A.~Vladu.
\newblock Towards deep learning models resistant to adversarial examples.
\newblock {\em arXiv preprint arXiv:1706.06083}, 2017.

\bibitem{moosavi2016universal}
S.-M. Moosavi-Dezfooli, A.~Fawzi, O.~Fawzi, and P.~Frossard.
\newblock Universal adversarial perturbations.
\newblock {\em arXiv preprint arXiv:1610.08401}, 2016.

\bibitem{moosavi2016deepfool}
S.-M. Moosavi-Dezfooli, A.~Fawzi, and P.~Frossard.
\newblock Deepfool: a simple and accurate method to fool deep neural networks.
\newblock In {\em Proceedings of the IEEE Conference on Computer Vision and
  Pattern Recognition}, pages 2574--2582, 2016.

\bibitem{papernot2016limitations}
N.~Papernot, P.~McDaniel, S.~Jha, M.~Fredrikson, Z.~B. Celik, and A.~Swami.
\newblock The limitations of deep learning in adversarial settings.
\newblock In {\em Security and Privacy (EuroS\&P), 2016 IEEE European Symposium
  on}, pages 372--387. IEEE, 2016.

\bibitem{papernot2016distillation}
N.~Papernot, P.~McDaniel, X.~Wu, S.~Jha, and A.~Swami.
\newblock Distillation as a defense to adversarial perturbations against deep
  neural networks.
\newblock In {\em Security and Privacy (SP), 2016 IEEE Symposium on}, pages
  582--597. IEEE, 2016.

\bibitem{sharif2016accessorize}
M.~Sharif, S.~Bhagavatula, L.~Bauer, and M.~K. Reiter.
\newblock Accessorize to a crime: Real and stealthy attacks on state-of-the-art
  face recognition.
\newblock In {\em Proceedings of the 2016 ACM SIGSAC Conference on Computer and
  Communications Security}, pages 1528--1540. ACM, 2016.

\bibitem{chen2017madry}
Y.~Sharma and P.-Y. Chen.
\newblock Breaking the {M}adry {D}efense model with {L}1-based adversarial
  examples.
\newblock {\em arXiv preprint arXiv:1710.10733}, 2017.

\bibitem{Szegedy14}
C.~Szegedy, W.~Zaremba, I.~Sutskever, J.~Bruna, D.~Erhan, I.~Goodfellow, and
  R.~Fergus.
\newblock Intriguing properties of neural networks.
\newblock In {\em International Conference on Learning Representations}, 2014.

\bibitem{tramer2017ensemble}
F.~Tram{\'e}r, A.~Kurakin, N.~Papernot, D.~Boneh, and P.~McDaniel.
\newblock Ensemble adversarial training: Attacks and defenses.
\newblock {\em arXiv preprint arXiv:1705.07204}, 2017.

\end{thebibliography}
\end{document}